\documentclass[letterpaper]{article} 
\usepackage{aaai2026}  
\usepackage{times}  
\usepackage{helvet}  
\usepackage{courier}  
\usepackage[hyphens]{url}  
\usepackage{graphicx} 
\usepackage{natbib}  
\usepackage{caption} 
\usepackage{subfigure}

\usepackage{booktabs}
\usepackage{makecell}
\usepackage{algorithm}
\usepackage{algorithmic}
\usepackage{amsmath}
\usepackage{newfloat}
\usepackage{listings}
\DeclareCaptionStyle{ruled}{labelfont=normalfont,labelsep=colon,strut=off} 
\floatstyle{ruled}
\newfloat{listing}{tb}{lst}{}
\floatname{listing}{Listing}
\title{STEM: Efficient Relative Capability Evaluation of LLMs\\through Structured Transition Samples}

\author {
    Haiquan Hu\textsuperscript{\rm 1},
    Jiazhi Jiang\textsuperscript{\rm 1},
    Shiyou Xu\textsuperscript{\rm 1},
    Ruhan Zeng\textsuperscript{\rm 1},
    Tian Wang\textsuperscript{\rm 1,*}
}
\affiliations {
    \textsuperscript{\rm 1}Institute of Artificial Intelligence and Future Networks, Beijing Normal University\\
    huhq99@yeah.net, jiangjz@bnu.edu.cn, 202311079173@mail.bnu.edu.cn, 202311109032@mail.bnu.edu.cn, cs\_tianwang@163.com
}

\usepackage{bibentry}

\begin{document}

\maketitle

\begin{abstract}
Evaluating large language models (LLMs) has become increasingly challenging as model capabilities advance rapidly. While recent models often achieve higher scores on standard benchmarks, these improvements do not consistently reflect enhanced real-world reasoning capabilities. Moreover, widespread overfitting to public benchmarks and the high computational cost of full evaluations have made it both expensive and less effective to distinguish meaningful differences between models. To address these challenges, we propose the \textbf{S}tructured \textbf{T}ransition \textbf{E}valuation \textbf{M}ethod (STEM), a lightweight and interpretable evaluation framework for efficiently estimating the relative capabilities of LLMs. STEM identifies \textit{significant transition samples} (STS) by analyzing consistent performance transitions among LLMs of the same architecture but varying parameter scales. These samples enable STEM to effectively estimate the capability position of an unknown model. Qwen3 model family is applied to construct the STS pool on six diverse and representative benchmarks. To assess generalizability. Experimental results indicate that STEM reliably captures performance trends, aligns with ground-truth rankings of model capability. These findings highlight STEM as a practical and scalable method for fine-grained, architecture-agnostic evaluation of LLMs.
\end{abstract}



\section{Introduction}

Large language models (LLMs) have made remarkable advancements in recent years, frequently achieving state-of-the-art (SOTA) performance on widely used benchmarks such as MMLU~\cite{Hendrycks21mmlu}, GPQA~\cite{Rein23gpqa}, GSM8K~\cite{Cobbe21gsm8k} and MATH~\cite{Hendrycks21math}. These benchmarks have become widely adopted standards for evaluating model performance in both academic research and public reporting~\cite{Chang24}. However, there is growing skepticism regarding the reliability and interpretability of these evaluation results~\cite{Xu24}. Despite impressive benchmark scores, user-reported experiences indicate a noticeable discrepancy between claimed accuracy and actual reasoning capability of LLMs~\cite{Balloccu24}. 

This growing gap has raised concerns regarding the reliability of current evaluation metrics~\cite{Ni25}. One key issue is that some benchmarks may be partially memorized by models during pretraining or alignment, leading to inflated scores that do not indicate actual reasoning capabilities~\cite{Hidayat25}. A notable example comes from the Qwen3 technical report~\cite{qwen3technicalreport}. As shown in Table~\ref{table1}, most benchmark results follow the expected trend where model performance scales positively with parameter size. However, this trend is clearly violated on the GPQA benchmark, where models ranging from Qwen3-8B and Qwen3-235B-A22B show irregular or even declining performance despite the increase in parameter size. Similar deviations are observed on GSM8K and MATH, indicating that scaling up model size does not consistently lead to better evaluation performance. These anomalies suggest that current evaluation methods may confuse real reasoning improvements with data overfitting~\cite{Zhang24}. Consequently, relying solely on aggregate accuracy across benchmarks can obscure a model’s actual reasoning boundaries which can lead to potentially misleading assessments. 

\begin{table*}[t]
\centering
\begin{tabular}{lccccccc}
\toprule
\textbf{Benchmarks} & \makecell[c]{MMLU\\(5-shot)} & \makecell[c]{MMLU-Pro\\(5-shot, CoT)} & \makecell[c]{SuperGPQA\\(5-shot, CoT)} & \makecell[c]{GPQA\\(5-shot, CoT)} & \makecell[c]{GSM8K\\(4-shot, CoT)} & \makecell[c]{MATH\\(4-shot, CoT)}\\ \midrule
Qwen3-235B-A22B & \textbf{87.81} & \textbf{68.18} & \textbf{44.06} & \underline{47.47} & \textbf{94.39} & \textbf{71.84}\\
Qwen3-32B       & \underline{83.61} & \underline{65.54} & \underline{39.78} & \textbf{49.49} & \underline{93.40} & 61.62\\
Qwen3-30B-A3B   & 81.38 & 61.49 & 35.72 & 43.94 & 91.81 & 59.04\\
Qwen3-14B       & 81.05 & 61.03 & 34.27 & 39.90 & 92.49 & \underline{62.02}\\
Qwen3-8B        & 76.89 & 56.73 & 31.64 & 44.44 & 89.84 & 60.80\\
Qwen3-4B        & 72.99 & 50.58 & 28.43 & 36.87 & 87.79 & 54.10\\
Qwen3-1.7B      & 62.63 & 36.76 & 20.92 & 28.28 & 75.44 & 43.50\\
Qwen3-0.6B      & 52.81 & 24.74 & 15.03 & 26.77 & 59.59 & 32.44\\ 
\midrule
LLaMA3-8B       & \makecell[c]{68.40\\(5-shot)} & -	& -	& \makecell[c]{34.20\\(0-shot)} & \makecell[c]{79.60\\(8-shot, CoT)} & \makecell[c]{30.00\\(4-shot, CoT)} \\
GLM4-9B         & 74.70 ( - ) & -	& -	& 34.30 ( - ) & 84.00 ( - ) & 30.40 ( - ) \\
\bottomrule
\end{tabular}
\caption{Evaluation scores of various LLMs on six benchmarks as reported in official technical reports. The highest and second-best scores are shown in bold and underlined, respectively. The symbol ‘-’ means missing information.}
\label{table1}
\end{table*}

Moreover, a deeper challenge lies in the \textbf{structural biases} present in many benchmarks~\cite{Majdinasab25}. Despite their widespread use, several popular benchmarks contain samples that are either too easy or too hard for models across a wide range of capabilities, making them insensitive to gradual improvements and limiting their capability to distinguish between models. Ideally, benchmark samples should cover various difficulty levels and show measurable performance transition as model scale grows, in accordance with scaling laws~\cite{kaplan20scalinglaws}. However, our statistical analysis of Qwen3 model family's performance in Table~\ref{table1} revealed that some benchmarks yield large deviations from expected trends, suggesting an imbalance in sample difficulty distribution. Appendix~\ref{Appendix:A} provides detailed statistical evidence. Multiple benchmarks contain a low proportion of intermediate sample that are informative for LLMs differentiation (e.g., only 20.07\% in GPQA and 34.80\% in GSM8K). Moreover, residual-based analyses reveal systematic departures from the scaling trend, where overabundance of simple and difficult samples leads to residual patterns that significantly deviate from linear regression predictions based on scaling laws. Additional metrics, such as standard deviation, skewness and kurtosis of residuals highlight imbalance in difficulty distribution.

To address these limitations, the \textbf{Structured Transition Evaluation Method (STEM)} is proposed as an efficient evaluation framework for estimating model capability using a small and carefully selected subset of high-discrimination samples, which vary distinctly in outcomes across models of different capacities and reflect capability boundaries. The STEM is motivated by the empirical observation that model performance tends to exhibit clear transitions as model scales increase. To systematically capture such phenomenon, the \textbf{Transition Index (TI)} is introduced to represent the smallest model that consistently answers a sample correctly. It allows us to identify a distinct subset of \textbf{Significant Transition Sample (STS)}, consisting of samples that show a clear transition from incorrect to correct responses along the model scale. A balanced selection of STS across TIs enables efficient and interpretable capability assessment without relying on full-benchmark evaluation. The STEM is validated across six benchmarks within Qwen3 model family, spanning from 0.6B to 235B parameters. Its effectiveness is further demonstrated by accurately positioning and distinguishing LLaMA3-8B~\cite{llama3modelcard} and GLM4-9B~\cite{glm2024chatglm} based on their performance within the Qwen3 capability range.

Our contributions are as follows:
\begin{itemize}
    \item A structural bias in widely used benchmarks is revealed, where the imbalanced distribution of samples across difficulty levels obscures differences in model capabilities.
    \item The \textit{significant transition sample} is defined and extracted from standard benchmarks to isolate samples without data contamination and enable interpretable model relative evaluation.
    \item The \textit{transition index} is introduced as a principled metric to capture capability transitions across model scales and classify high-discrimination STS.
    \item The STEM framework is proposed and validated, demonstrating that a small-STS set can efficiently achieve relative evaluation of LLMs capability within Qwen3 model family and distinguish models with similar aggregate scores, such as LLaMA3-8B and GLM4-9B.
\end{itemize}

\section{Related Work}

\subsection{LLM Evaluation Paradigms}
Evaluating LLMs has become a central concern as their capabilities rapidly expand. Current evaluation methods generally fall into two mainstream categories: full-benchmark evaluation and random-sample evaluation. The former involves testing the LLM on entire standard benchmark such as MMLU~\cite{Hendrycks21mmlu}, GPQA~\cite{Rein23gpqa}, GSM8K~\cite{Cobbe21gsm8k} and MATH~\cite{Hendrycks21math}. It provides stable and comprehensive results but incurs high computational costs, making it often impractical for routine assessments~\cite{Zhang25,Biderman24}. Random-sample evaluation, on the other hand, offers a more resource-efficient alternative by selecting subsets of benchmark items for testing, typically in multiple sampling rounds. While this method reduces computational burden, it raises concerns about representativeness, reproducibility and sensitivity to model differences, especially when the benchmark contains a high proportion of trivial or low-difficulty items~\cite{Madaan24}. These shortcomings become particularly noticeable when distinguishing between models with similar performance levels.

\subsection{Benchmark Structural Bias}
Recent studies have increasingly highlighted structural bias in public benchmarks for LLMs. Data contamination, a type of structural bias, has become a central concern in the evaluation of LLMs. Studies have demonstrated that LLMs often memorize benchmark samples during pretraining or alignment stages, resulting in artificially inflated performance that fails to reflect their true reasoning abilities~\cite{Zhou23,López25}. To assess the extent of such leakage, several detection methods have been proposed, including n-gram-based method for GSM8K and MATH~\cite{Xu24permutation}, as well as permutation~\cite{Ni25} and semi-half techniques for multiple-choice benchmarks~\cite{Hidayat25}. However, these methods are typically designed for specific types of benchmarks and may not generalize well across different evaluation settings. The STS defined in this paper provides a sample-level data contamination analysis tool and reveals a previously underexplored structural bias, that is the imbalance in sample difficulty distribution.

\subsection{Emergent Abilities of LLMs}
Emergent abilities in large language models have been widely recognized, referring to the sudden appearance of complex capabilities as model scale increases~\cite{Wei22}. Most existing studies concentrate on task-level emergence, where improvements are observed across entire benchmarks~\cite{Schaeffer23,Lu24}. However, the hypothesis that certain individual samples may display sensitivity to model scale has not been thoroughly investigated. Our work extends this direction by introducing a structured evaluation perspective based on capability transition patterns observed across model scales.

\section{Methodology}

\subsection{Overview of STEM}
The advancement of LLMs has outpaced the scalability of traditional evaluation methods. For this reason, we propose Structured Transition Evaluation Method (STEM), a lightweight LLMs evaluation framework designed to quickly and reliably estimate a model’s relative capability within a known model family.

\begin{figure*}[t]
    \centering
    \includegraphics[width=0.95\linewidth]{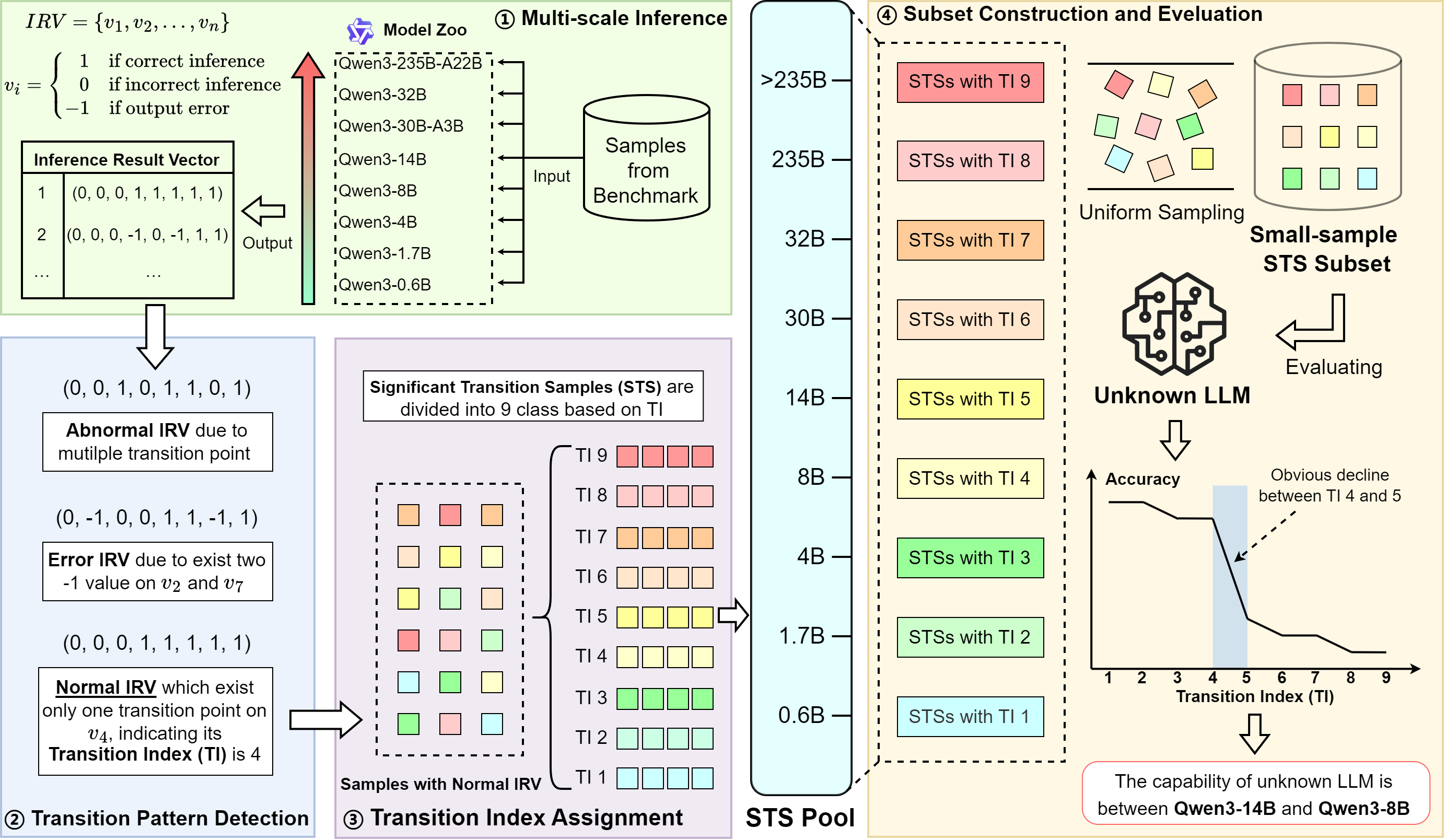}
    \caption{The framework of STEM. The first three steps are an offline preparatory process to build the STS pool, where samples are screened and categorized into 9 difficulty levels (TI 1 to 9). The fourth step outlines the evaluation process, where a small and difficulty-uniform subset is constructed by sampling from the pre-built STS pool to efficiently assess an unknown LLM.}
    \label{fig1}
\end{figure*}

The core idea of STEM is based on the observed consistency and monotonicity of model's performance improvement as parameter size increases. For a given dataset, most samples indicate a predictable capability transitions. If a small model answers a question correctly, larger models tend to do so as well. Conversely, there exists a subset of samples for which only sufficiently powerful models can provide correct answers. As shown Figure~\ref{fig1}, STEM identifies such samples and leverages them as reference to evaluate unknown models. Specifically, the detailed steps are as follows:
\begin{enumerate}
    \item \textbf{Multi-scale Inference}: Benchmark evaluations are conducted across a series of same-architecture LLMs with increasing parameter sizes (e.g., 0.6B to 235B). The inference results for each sample consist of a ternary vector.
    \item \textbf{Transition Pattern Detection}: Based on the inference result vectors, a set of samples is identified in which a single 0-to-1 transition occurs. In these cases, all models below a certain size answer incorrectly, while all larger models answer correctly. These significant transition samples serve as indicators of capability boundaries across different model scales.
    \item \textbf{Transition Index Assignment}: Each STS is assigned a Transition Index reflecting the model size where its answer changes from incorrect to correct, forming the basis for a structured classification according to difficulty.
    \item \textbf{Subset Construction and Evaluation}: A small and balanced subset of STS is constructed by selecting an equal number of samples from each transition index level. When evaluating an unknown model, its performance on these structured subsets can be used to infer its capability range within the known model family.
\end{enumerate}

These components form a coherent and efficient evaluation process. Multi-scale inference produces result vectors that identify performance transition patterns. These patterns are then encoded as transition indexes, allowing samples to be systematically categorized by difficulty. A one-time extraction of significant transition samples from the full benchmark yields a pool of high-discrimination samples, which serves as a reusable basis for evaluation. For subsequent capability assessments, only a small and balanced subset needs to be sampled from the fixed STS pool, supporting efficient, interpretable and architecture-agnostic capability estimation. 

\subsection{Inference Result Vector}
To enable a systematic analysis of how individual samples respond across models with increasing capability, an Inference Result Vector (IRV) is defined for each data sample. The IRV is a ternary vector denoted as $IRV=\{v_1,v_2,\dots,v_n\}$, where $v_i\in \{-1,0,1\}$ corresponds to the inference outcome of the $i$-th model in a fixed and ordered model series. Specifically, $v_i=1$ indicates a correct answer, $v_i=0$ indicates an incorrect but well-formed answer, $v_i=-1$ indicates a failure to produce a valid output, such as formatting errors or instruction-following issues. The model sequence is denoted as $M_1$, $M_2$, ..., $M_n$, where each model $M_i$ is progressively more capable with increasing parameter size or architectural complexity. The index order is strictly monotonic with respect to model scale, such that $M_i\prec M_{i+1}$ reflects increasing model capability. 

The IRV provides a unified representation of the performance of models with different capabilities on a single sample, enabling analysis of whether it follows expected scale law on LLMs. Ideally, IRVs should exhibit a single and monotonic transition from incorrect to correct responses, indicating a clear model capability threshold. Departures from this idealized pattern, such as multiple fluctuations, may indicate memorization effects of LLMs, signaling potential overfitting rather than genuine generalization. Consequently, IRV not only facilitate to classify significant transition sample and calculate transition index, but also function as a diagnostic tool for identifying sample with data contamination. 

\subsection{Significant Transition Sample}
Building on the inference result vector, Significant Transition Samples (STS) are defined as those exhibit single 0-to-1 transition from incorrect to correct responses across an ordered sequence of models. Formally, a data sample qualifies as an STS if its IRV satisfies the following conditions:
\begin{itemize}
    \item \textbf{Monotonicity}: For all model index $i<k$, the IRV satisfy $v_i=0$, and all $i^{'}>k$, $v_{i^{'}}=1$, which ensures that smaller models consistently fail to infer the sample, while larger models always succeed.
    \item \textbf{Uniqueness}: The IRV must contain only one transition point from 0 to 1, with no reversals in subsequent model outputs. Such property eliminates ambiguous patterns caused by overfitting.
\end{itemize}
Each STS is associated with a transition index $k$, denoting the smallest model capability required to solve the sample reliably. For instance, an $IRV=(0,0,0,1,1,1,1,1)$ indicates a transition at index 3, implying that the fourth model in the sequence is the first to handle task successfully and stably. However, the abnormal IRV resembling $(0,0,1,0,1,1,0,1)$ indicates that data contamination caused by overfitting during LLM training. Moreover, the erroneous IRV resembling $(0,0,0,-1,0,-1,1,1)$ means that samples respond to errors in multi-scale inference.

Such samples serve as effective capability evaluation, providing discrete decision boundaries that distinguish the competence levels between adjacent model scales. By filtering out non-monotonic and noisy inference result vector, the STS set improves robustness in model capability evaluation and reduces the impact of data contamination from memorized samples.

\subsection{Structured Evaluation Protocol}
To support efficient and interpretable estimation of an unknown model's capability, a structured evaluation protocol is designed, leveraging the distribution of STS across model scales. The evaluation proceeds as follows: 
\begin{enumerate}
    \item \textbf{STS Pool Construction}: For each benchmark, the full set of STS samples is extracted using the rules described in Section 3.3. Each sample is tagged with a transition index $k\in\{1,2,\dots ,n+1\}$, where $k$ denotes the index of the smallest model that consistently produces a correct answer. $k=n+1$ indicates that the corresponding sample cannot be correctly answered by the largest model.
    \item \textbf{Balanced Subset Sampling}: To construct a structural and representative test set, an equal number of STS samples is randomly selected from each transition index class. The test set coverages across all capability thresholds while limiting the total number of sample, ensuring both lightweight and interpretability.
    \item \textbf{Target Model Capability Evaluation}: The unknown model is evaluated on a balanced STS subset, with binary correctness labels recorded. Its capability boundary is defined as the lowest transition index where accuracy remains consistently high. A sharp drop in accuracy at higher indices indicates the model’s capability lies between the highest index it handles reliably and the point where performance begins to decline, allowing inference of its relative position among reference models.
\end{enumerate}

\section{Experiment Design and Analysis}

\subsection{Experiment settings}
To validate the effectiveness and generalizability of the proposed Structured Transition Evaluation Method (STEM), extensive experiments are conducted across a range of state-of-the-art LLMs and widely adopted benchmarks.

The validation relies on a \textit{reference model family} to construct the STS pool, which is a critical component of the STEM framework. This process requires a series of same-architecture models across a wide range of parameter sizes. The \textbf{Qwen3 model family}~\cite{qwen3technicalreport} was selected as it uniquely satisfies these criteria, offering eight models from 0.6B to 235B parameters (e.g., Qwen3-0.6B, Qwen3-1.7B, Qwen3-4B, Qwen3-8B, Qwen3-14B, Qwen3-30B-A3B, Qwen3-32B and Qwen3-235B-A22B). In contrast, many contemporary open-source models like LLaMA3 are released in only two parameter sizes, which is insufficient for our fine-grained analysis. While older families such as OPT~\cite{zhang22opt} or LLaMA2~\cite{touvron23llama2} offer more size variations, their performance is no longer representative of the current state-of-the-art. The practical scarcity of suitable, modern, and scale-controlled reference models makes Qwen3 the most viable choice.

To test the transferability and architecture-agnostic properties of STEM, the validation also needs \textit{external models}. For this purpose, \textbf{LLaMA3-8B}~\cite{llama3modelcard} and \textbf{GLM4-9B}~\cite{glm2024chatglm} were selected. Despite being trained independently on different architectures, the difference in capability between them is minimal. They serve as excellent test cases for exhibiting that the STEM framework can be effectively transferred to assess models from entirely different architectures.

A comprehensive validation is conducted on six widely-adopted benchmarks, which cover a wide range of difficulty levels and task types, including general reasoning of \textbf{MMLU}~\cite{Hendrycks21mmlu} and \textbf{MMLU-Pro}~\cite{wang2024mmlupro}, domain-specific question answering of \textbf{GPQA}~\cite{Rein23gpqa} and \textbf{SuperGPQA}~\cite{pteam2025supergpqa}, and mathematic reasoning of \textbf{GSM8K}~\cite{Cobbe21gsm8k} and \textbf{Math}~\cite{Hendrycks21math}. These diversity are central to analyzing the discernibility of STS and confirming the robustness of the STEM framework across different capabilities.



To assess the relative efficiency and effectiveness, three evaluation strategies are compared:
\begin{itemize}
    \item \textbf{Random sampling}: For each benchmark, multiple random subsets of equal size are sampled for evaluation. The final accuracy is averaged across several runs. It reduces cost but suffers from sample variance and limited discernibility for LLMs capability.
    \item \textbf{Bayesian}~\cite{xiao25}: Leverages a manually selected query set and a series of pre-evaluated anchor models. By formalizing model ranking as a Bayesian hypothesis testing problem, a probabilistic distribution of model capability across range defined by anchors.
    \item \textbf{STEM (Ours)}: A fixed-size subset is constructed from STS using balanced sampling based on the transition index. Each subset contains multiple samples per difficulty level, which are utilized to infer the model’s capability range within the reference model family.
\end{itemize}

In all experiments, a consistent prompting format is applied across all LLMs. Except for the Qwen3-235B-A22B, Qwen3-32B and Qwen3-30B-A3B, which are accessed via an official API service, all other models are deployed locally using the vLLM framework~\cite{kwon23vllm} with FP32 precision on a single NVIDIA A100 GPU. To ensure that LLMs reflect their native capabilities, all experiments employ a zero-shot and non-CoT strategy.

\subsection{Model capability measurement based on official technical reports} 
To establish a reference standard for ranking existing LLMs, an unified and consistent measurement of LLM capability is constructed based on publicly available official technical reports. Considering the sample difficulty distribution of benchmarks, simply averaging raw scores would fail to capture their relative informativeness in model comparison.

To this end, the discriminability $D_j$ of $j$-th benchmark is designed as Equation~\ref{eqn:1}, which reflects both its capacity to differentiate model capabilities and its alignment with the scaling laws. Given the parameter sizes of LLMs family $P$ and their scores on $j$-th benchmark $S_j$:
\begin{equation}
\label{eqn:1}
    D_j=\sigma_{S_j}\times \rho_{S_j,\log (P)} 
\end{equation}
where $\sigma_{S_j}$ is the standard deviation of LLMs' scores on $j$-th benchmark, displaying the discernibility of performance gap between LLMs. $\rho_{S_j,\log (P)}$ is the Pearson Correlation Coefficient between the logarithm of model parameter size $\log (P)$ and the benchmark scores $S_j$, which indicates the consistency of model performance with scaling laws. Based on the discriminability $D_j$ shown in Table~\ref{table2}, the weight $w_j$ of $j$-th benchmark score is calculated as Equation~\ref{eqn:2}:
\begin{equation}
\label{eqn:2}
    w_j=\frac{D_j}{\sum_{j=1}^{m} D_j}
\end{equation}

\begin{table}[h]
\setlength{\tabcolsep}{2pt}
\centering
\small
\begin{tabular}{cccccccc}
\toprule
 & MMLU    & MMLU-Pro & SuperGPQA & GPQA & GSM8K & MATH \\ 
\midrule
$D$    & 10.36 & 13.13 & 8.75 & 7.04 & 9.57  & 10.77 \\
\bottomrule
\end{tabular}
\caption{The discernibility value of six benchmarks}
\label{table2}
\end{table}
The reference scores of LLMs aggregate benchmark scores according to their weight $w_j$. Considering the absence of LLaMA3-8B and GLM4-9B scores on MMLU-Pro and SuperGPQA, the reference score is calculated based only on MMLU, GPQA, GSM8K, and MATH to determine their relative capability within the Qwen3 model family. 

\textbf{Reference ranking}: From the perspective of capability shown in Table~\ref{table3}, Qwen3-4B $<$ GLM4-9B $<$ Qwen3-1.7B, while Qwen3-1.7B $<$ LLaMA3-8B $<$ Qwen3-0.6B.

\begin{table}[ht]
\centering
\begin{tabular}{l|c||l|c}
\toprule
LLMs & Scores & LLMs & Scores \\
\midrule
Qwen3-235B-A22B & 77.39 & Qwen3-4B   & 64.61 \\
Qwen3-32B       & 73.45 & \textbf{GLM4-9B} & \textbf{56.88} \\
Qwen3-30B-A3B   & 70.66 & Qwen3-1.7B & 54.01 \\
Qwen3-14B       & 70.84 & \textbf{LLaMA3-8B} & \textbf{53.90} \\
Qwen3-8B        & 69.53 & Qwen3-0.6B & 43.86 \\ 
\bottomrule
\end{tabular}
\caption{The reference ranking of LLMs across MMLU, GPQA, GSM8K and MATH}
\label{table3}
\end{table}

\subsection{Sample-level analysis of data contamination}

To ensure that the evaluation of LLMs is based on samples with genuine discernibility, this section aims to identify and filter out potentially contaminated data. By analyzing inference consistency across Qwen3 model family, samples exhibiting irregular or non-monotonic behavior are tagged as abnormal, allowing a cleaner and more trustworthy subset to be constructed for following evaluation.

As introduced in Section 3.2, each sample is associated with an Inference Result Vector (IRV). A sample is considered valid if its IRV is monotonic non-decreasing and contains exactly one 0-to-1 transition. The samples that violate such pattern are marked as \textbf{abnormal samples}, indicating a higher likelihood of data contamination or memorization. It is based on the logical premise that a more capable model should not fail a question that a less capable one success. A non-monotonic IRV such as (0, 1, 0, ...) where a smaller model succeeds but a larger one fails, violates this assumption. Such inversion suggests the smaller model's success is likely due to memorizing a contaminated sample from its training data, rather than genuine reasoning, making the sample unreliable for assessing true capability growth.

\begin{figure}[ht]
    \centering
    \includegraphics[width=1.0\linewidth]{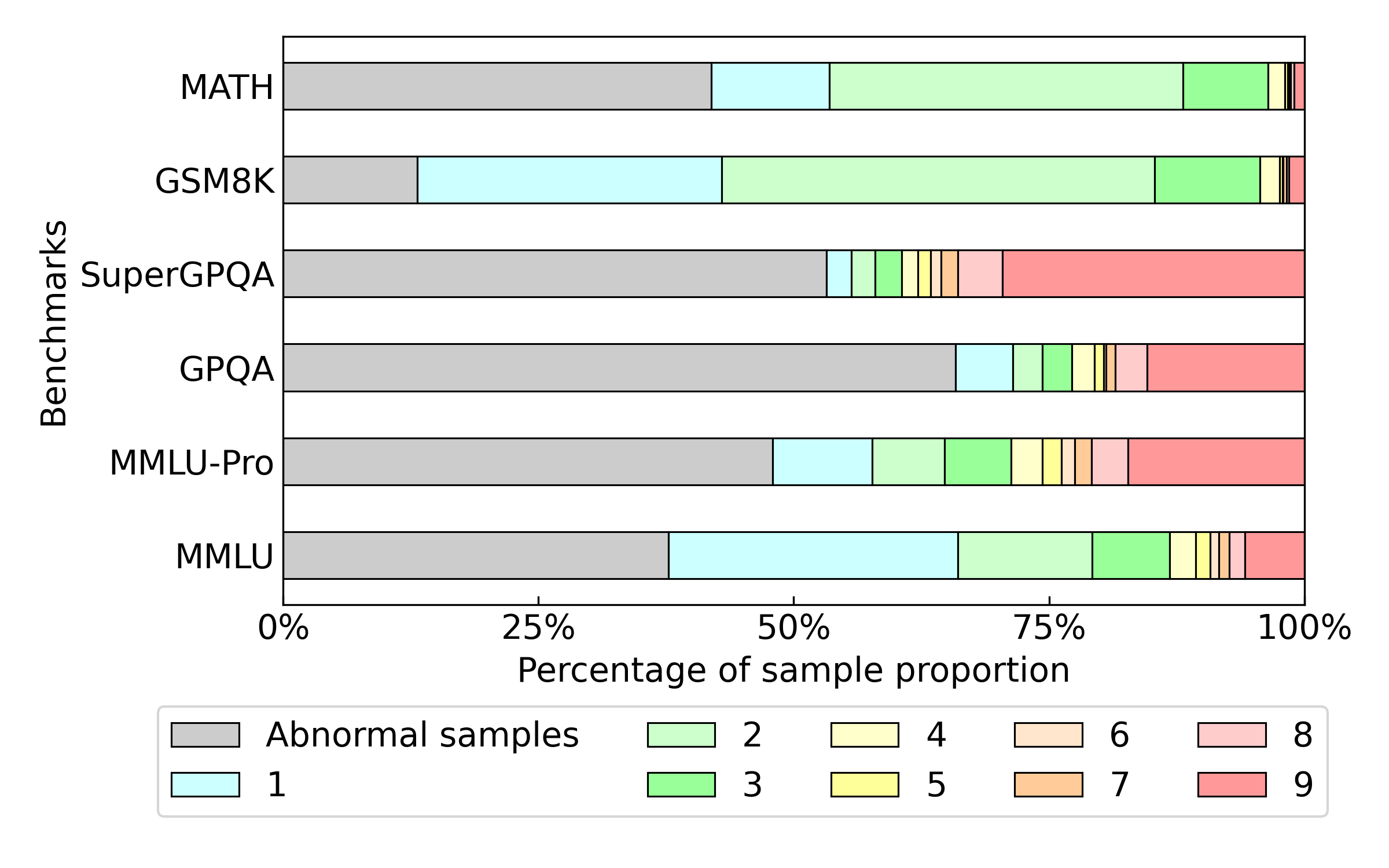}
    \caption{The proportion of sample with TI from 1 to 9 across six benchmarks}
    \label{fig2}
\end{figure}

Figure~\ref{fig2} shows the sample distribution based on transition index and the proportion of abnormal samples across the six benchmarks. The results indicate significant differences in the proportion of abnormal samples among different benchmarks. Specifically, about 65.85\% of the samples in GPQA do not meet the structural requirements of IRV, while the proportions of abnormal samples in SuperGPQA and MMLU-Pro are 53.20\% and 47.93\%, respectively. MMLU and MATH also have considerable proportions of abnormal samples, accounting for 37.72\% and 41.94\%, respectively. In contrast, GSM8K has the lowest proportion of abnormal samples at only 13.16\%, indicating that its samples better conform to the pattern of stable transitions as model capability improves.

The high proportion of abnormal samples highlights the importance of selecting STS and serves as evidence of structural biases in current benchmarks. Therefore, constructing evaluation subsets with reasonable structure and clear difficulty distribution is crucial for improving the reliability and interpretability of model evaluation.

\begin{figure}[ht]
    \centering
    \includegraphics[width=0.95\linewidth]{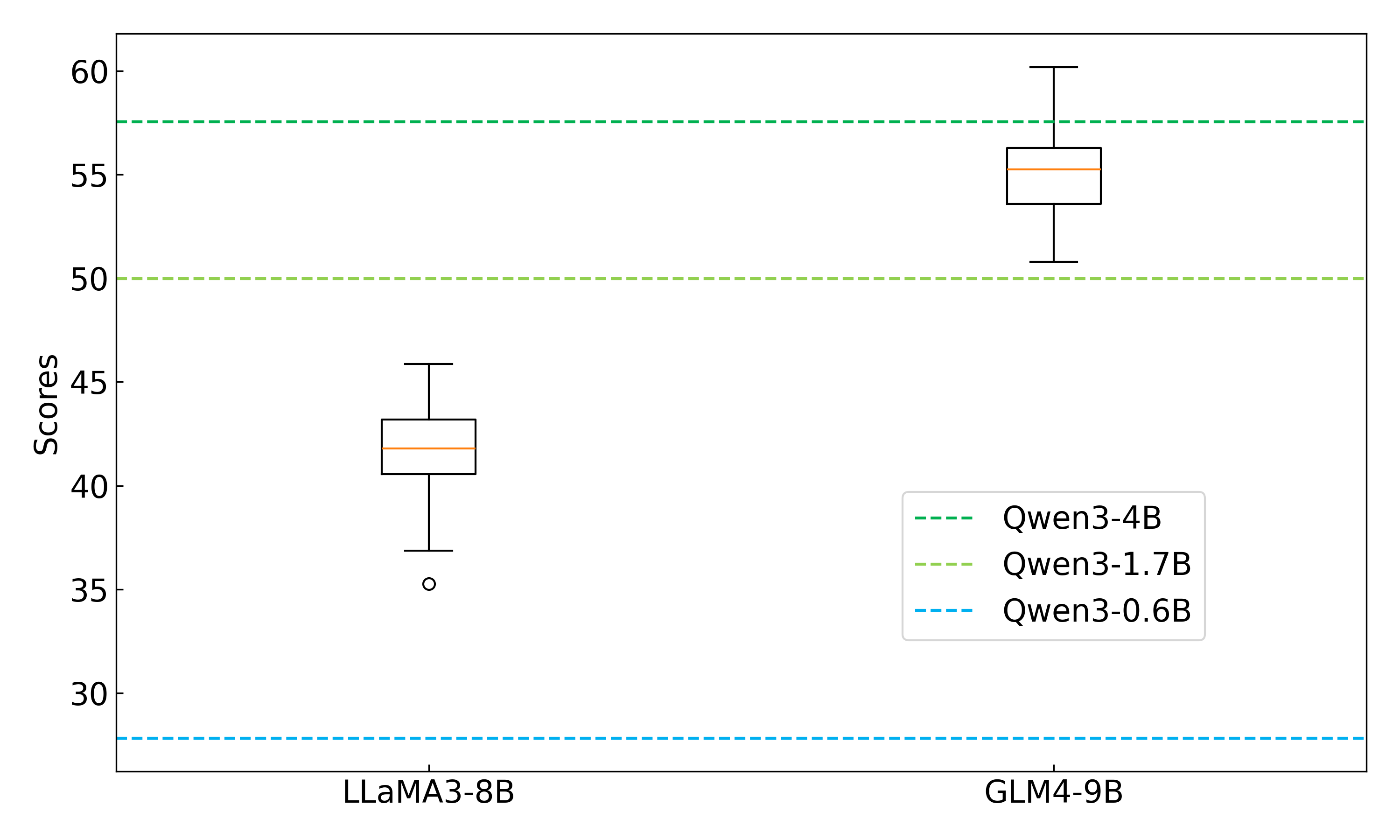}
    \caption{Random sampling estimation for LLaMA3-8B and GLM4-9B}
    \label{fig3}
\end{figure}

\subsection{Model capability evaluation}

To assess the effectiveness and reliability of the proposed STEM method, this section compares its performance against both random sampling and Bayesian evaluation methods. Each method was used to estimate the relative capability of LLaMA3-8B and GLM4-9B within the Qwen3 model family, with the full-benchmark results serving as the ground truth. The stability of the results was ensured by using a sample size of 100 and repeating each experiment 100 times.

\begin{figure}[ht]
    \centering
    \subfigure[LLaMA3-8B]{\includegraphics[width=0.48\linewidth]{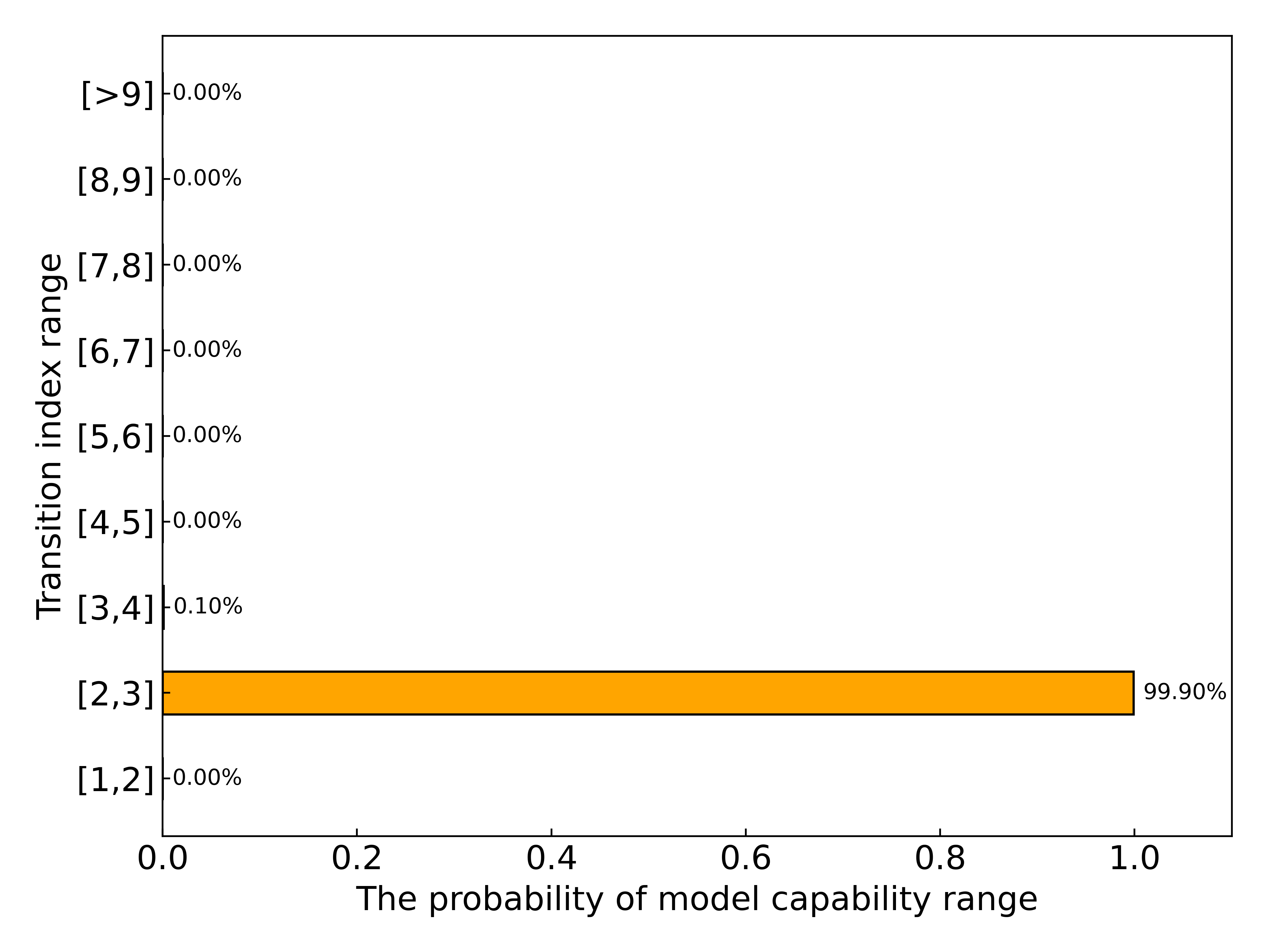}}
    \subfigure[GLM4-9B]{\includegraphics[width=0.48\linewidth]{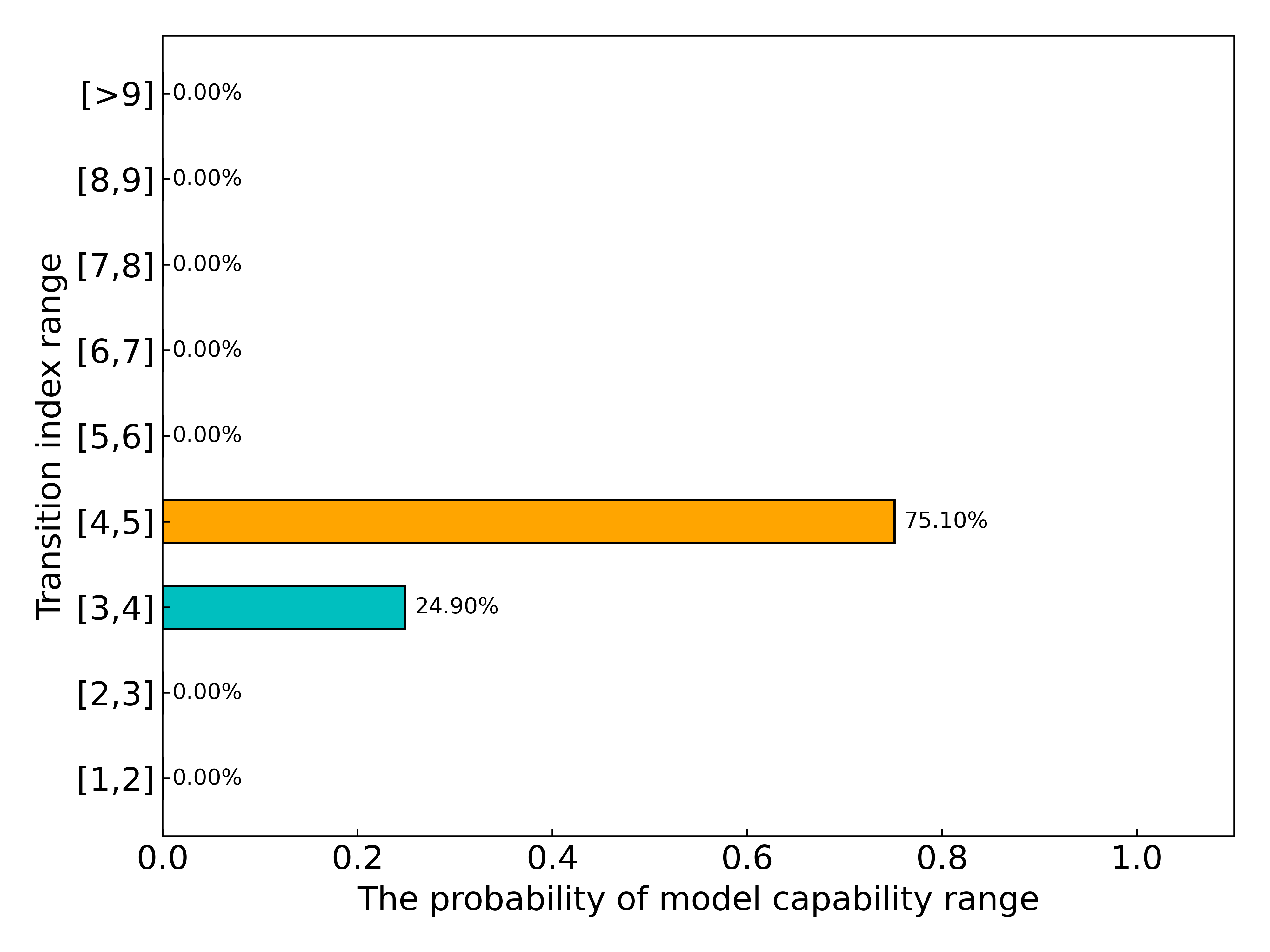}}
    \caption{The Bayesian estimation based on the probability of capability range for LLaMA3-8B and GLM4-9B}
    \label{fig4}
\end{figure}

The evaluation first employed the random sampling method, with results illustrated in Figure~\ref{fig3}. It positioned LLaMA3-8B and GLM4-9B with mean scores of 41.82 and 54.98, respectively. On average, the scores from random sampling are consistent with the ground-truth rankings presented in Appendix Table~\ref{tableB}. Specifically, this method correctly positions LLaMA3-8B's capability between that of Qwen3-0.6B and Qwen3-1.7B, and GLM4-9B's between Qwen3-1.7B and Qwen3-4B. However, the method's overall reliability is severely compromised by its high intrinsic variance. Due to its inherent stochasticity, evaluation scores for GLM4-9B occasionally surpassed the reference score of Qwen3-4B, which is led to an inaccurate estimation in 12\% of the trials, resulting in an 88\% accuracy rate for this model, as summarized in Table~\ref{table4}.

As shown in Figure~\ref{fig4}, the Bayesian method yields a probabilistic estimation of model capability. It assigned a 99.9\% probability that LLaMA3-8B's capability falls between Qwen3-1.7B and Qwen3-4B, and a 75.1\% probability that GLM4-9B lies between Qwen3-14B and Qwen3-8B. Although the method correctly identified that GLM4-9B is more capable than LLaMA3-8B, it systematically and significantly overestimated both models' capabilities compared to their ground-truth positions. Consequently, the Bayesian approach failed to correctly identify the true capability interval for either model in any of the repeated trials, leading to a 0\% accuracy rate.
 
\begin{figure}[ht]
    \centering
    \subfigure[LLaMA3-8B]{\includegraphics[width=0.48\linewidth]{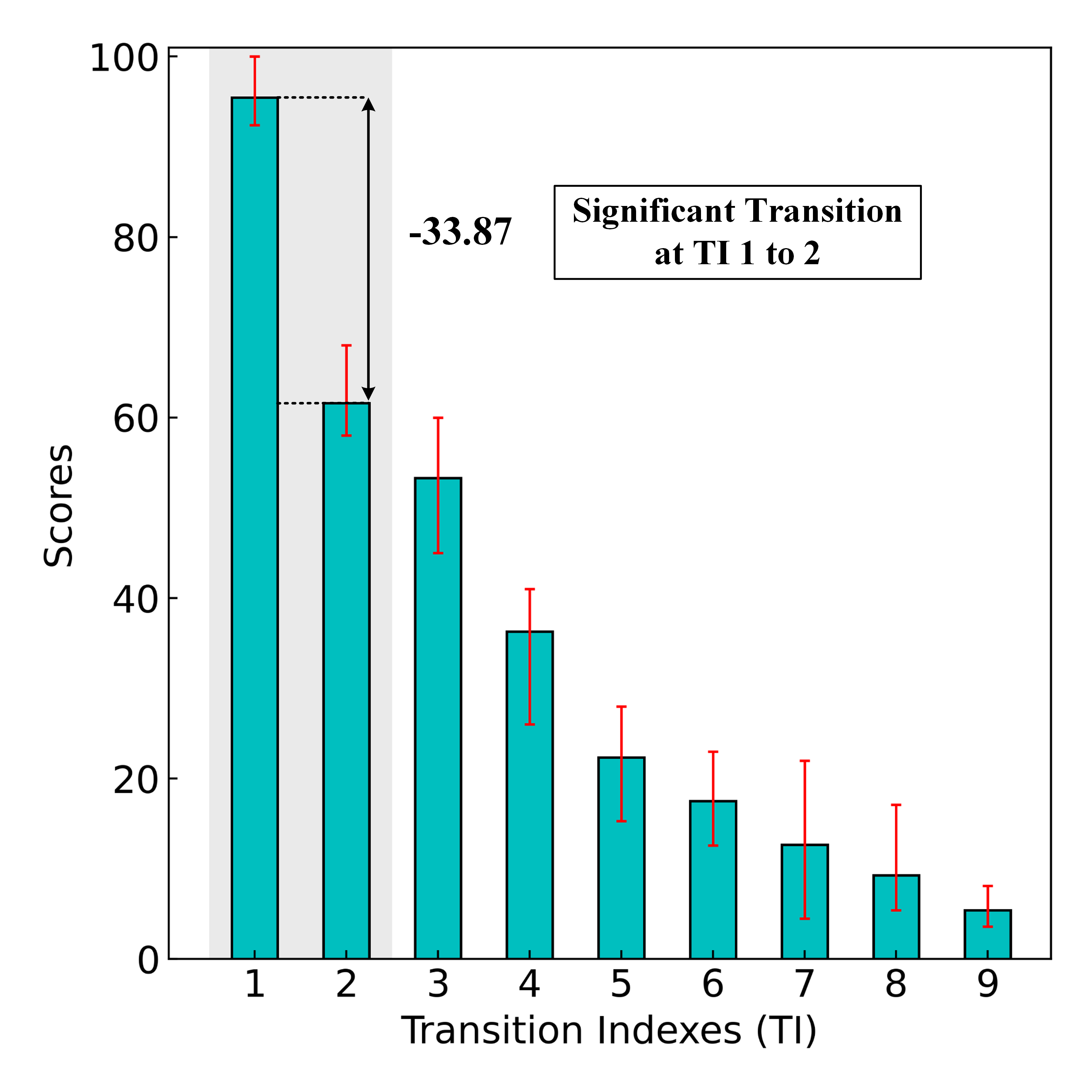}}
    \subfigure[GLM4-9B]{\includegraphics[width=0.48\linewidth]{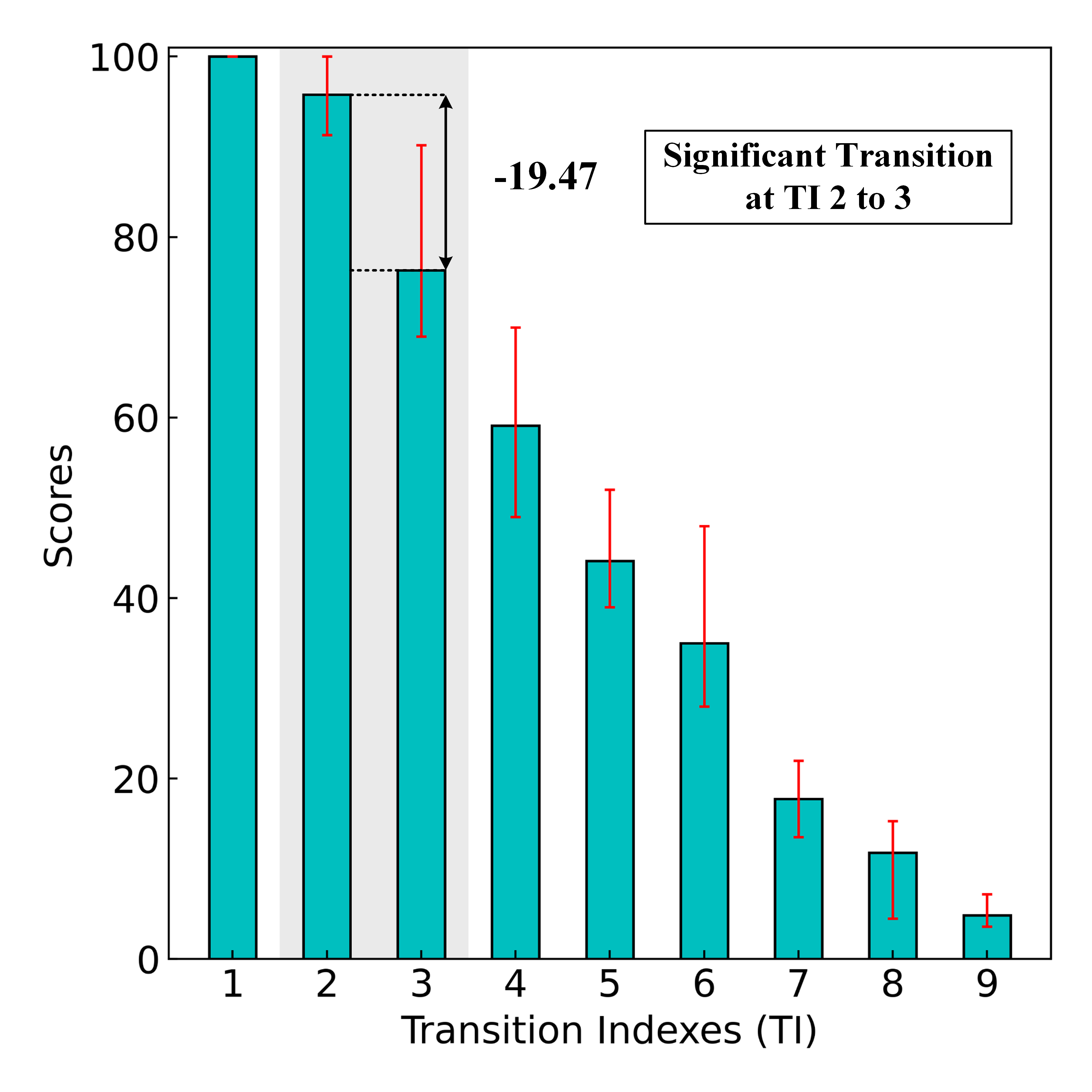}}
    \caption{The STEM estimation based on transition index for LLaMA3-8B and GLM4-9B}
    \label{fig5}
\end{figure}

For the STEM method, model capability is determined by identifying the first significant performance drop along the TIs. As shown in Figure~\ref{fig5}, LLaMA3-8B exhibits a sharp score decline of 33.87 between TI 1 and 2, indicating its capability lies between Qwen3-0.6B and Qwen3-1.7B. Similarly, GLM4-9B experiences a noticeable drop of nearly 20 points and increased variance between TI levels 2 and 3, suggesting its capability falls between Qwen3-1.7B and Qwen3-4B. These results are fully consistent with the ground-truth capability ranges of both models, demonstrating the effectiveness and precision of the STEM method.

\begin{table}[ht]
\centering
\begin{tabular}{lccc}
\toprule
LLMs & Random Sampling & Bayesian & STEM \\
\midrule
LLaMA3-8B & 100\% & 0\% & 100\% \\
GLM4-9B & 88\% & 0\% & 100\% \\
\bottomrule
\end{tabular}
\caption{The LLM capability estimation accuracy rate of three evaluation strategies.}
\label{table4}
\end{table}

The final accuracy rates of the three evaluation strategies, summarized in Table~\ref{table4}, highlight the superiority of the proposed framework. While random sampling suffers from unreliability due to high variance and the Bayesian method is hampered by systematic overestimation, the STEM method achieves a 100\% accuracy rate, correctly identifying the precise capability intervals for both LLaMA3-8B and GLM4-9B in all trials. This demonstrates that STEM is not only efficient but also significantly more reliable and precise than established baseline methods.

\section{Discussion}

\subsection{Failure of Bayesian Baseline}

The Bayesian evaluation~\cite{xiao25} method aligns with our core idea, which both seek to position an unknown model's capability by referencing a set of anchor models. However, there are three technological flaws, leading to 0\% accuracy in identifying capability intervals. 

Firstly, it relies on manually curated query samples, which is a subjective, labor-intensive, and fragile process, particularly ineffective in cross-domain scenarios. In contrast, the STEM adopts a data-driven approach, automatically extracting STSs based on performance transition across model scales, ensuring objectivity and adaptability. Secondly, the reference model hierarchy in Bayesian is empirical, using closed-source and various architecture LLMs (e.g., GPT-3.5-Turbo, GPT-4o). Such ambiguous capability gaps undermine interpretability. By contrast, STEM uses Qwen3 model family with consistent architecture and varying scale, enabling transparent, fine-grained capability tracking. Finally, the Bayesian method is fragile and imprecise. Its posterior estimates are highly sensitive to a few correct samples. Due to coarse-grained and opaque model anchors, this sensitivity can cause systematic overestimation. STEM’s dense, scale-controlled model hierarchy offers more reliable and granular evaluations, highlighting the Bayesian method’s unsuitability for nuanced capability assessment.

\subsection{Threats to Validity}

While the STEM framework demonstrates significant advantages, its findings are subject to several threats to validity that warrant consideration. A primary internal threat is the construct of "capability", which is intrinsically defined by performance on a specific set of benchmarks. This limits the interpretation of the measured construct to those domains. Furthermore, the framework’s internal logic rests on the critical assumption that STS transfer effectively across different model architectures. While supported by the current results, the universal applicability of this assumption is not guaranteed and remains a potential vulnerability. Regarding external validity, the most significant threat is the method's dependency on a scale-controlled reference model family. The scarcity of such families severely constrains the widespread applicability of the STEM framework. Moreover, the static nature of the resulting STS pool poses a practical challenge to long-term generalizability, as it requires periodic and computationally expensive recalibration to remain relevant in a rapidly advancing field.

\section{Conclusion}

This paper introduces STEM, a structured evaluation method designed to estimate LLM capabilities with minimal inference cost and maximal interpretability. By leveraging consistent performance transitions across model scales, STEM extracts high-discrimination samples and constructs balanced evaluation subsets based on Transition Index. Experiments across diverse benchmarks and model families show that STEM consistently produces accurate capability rankings and clear capability localization. Compared to existing alternatives such as random sampling and Bayesian estimation, STEM achieves better stability, clearer separation, and full correctness in model ordering. The results highlight STEM’s practical utility as a scalable, architecture-agnostic evaluation tool for rapidly evolving LLM ecosystems. Future work will explore extending the method to generative tasks and incorporating more robust detection of data contamination.

\bibliography{STEM}

\twocolumn[\clearpage]
\appendix
\section{Analysis of Sample Difficulty Distribution}
\label{Appendix:A}
Scaling laws provide an important mathematical framework for characterizing LLM capabilities, suggesting that performance should increase in a stable and log-linear relationship with parameter size. Consequently, a benchmark with a uniform or balance difficulty distribution should yield model scores that follow the expected scaling trend. The subsequent analyses are conducted under this assumption.

\begin{table*}[t]
\centering
\begin{tabular}{cccccccc}
\toprule
Sample Group & MMLU    & MMLU-Pro & SuperGPQA & GPQA & GSM8K & MATH \\ 
\midrule
Difficult    & 12.19\% & 31.82\% & 55.94\% & 52.53\% & 5.61\%  & 28.16\% \\
Intermediate & 35.00\% & 43.44\% & 29.03\% & 20.70\% & 34.80\% & 39.40\% \\
Simple       & 52.81\% & 24.74\% & 15.03\% & 26.77\% & 59.59\% & 32.44\% \\
\bottomrule
\end{tabular}
\caption{The proportions of sample distribution for six benchmarks}
\label{table:A1}
\end{table*}

At first, the samples are categorized into three groups: (1) \textbf{Simple}, correctly answered by the smallest model; (2) \textbf{Difficult}, not correctly answered even by the largest model; (3) \textbf{Intermediate}, correctly answered by one or more models within the Qwen3 model family. Table~\ref{table:A1} shows the sample proportions for the six benchmarks. From the perspective of distinguishing LLM capabilities, both simple and difficult samples are uninformative. Notably, MMLU and GSM8K are dominated by simple samples ($>$50\%), whereas GPQA and SuperGPQA contain a surplus of difficult samples ($>$50\%). 

\textbf{Preliminary Finding}: The difficulty distribution of the six benchmarks overall shows a trend of polarization.

To further probe these distributional biases, a scaling laws-based line regress fitting and residual-based analysis are conducted. The performance of LLMs with various parameter size $P=\{235,32,30,14,8,4,1.7,0.6\}$ is modeled using linear regression:
\begin{equation}
    S_i=\alpha \times \log (P_i)+\beta +\epsilon_i
\end{equation}
where $S_i$ indicates the $i$-th LLM's benchmark scores, $\log (P_i)$ is logarithm value of its parameter size, $\epsilon_i$ is residual item, calculated as:
\begin{equation}
    \epsilon_i=S_i-\hat{S_i}=S_i-(\alpha \times \log (P_i)+\beta)
\end{equation}

The results of regression are shown as Figure~\ref{fig:A1}, where the range of y-axis is uniformly set at 60 scores. Meanwhile, the corresponding residual items are detailed in Table~\ref{table:A2}. Based on the definition of residual, $\epsilon_i>0$ indicates LLM's performance is higher predicted value, implying the sample is simpler than expected; $\epsilon_i<0$ indicates LLM's performance is lower predicted value, implying the sample is more difficult than expected; $\epsilon_i=0$ indicates the LLM's performance and sample difficulty aligns with the expectations. 

Analysis of the residuals reveals a consistent trend: residuals are negative for both the smallest and the largest models, while those corresponding to medium-sized models are uniformly positive. This pattern is not indicative of dominance by medium-difficulty samples. Instead, it is a direct artifact of the sample difficulty polarization previously identified in Table~\ref{table:A1}. 

\begin{figure}[ht]
    \centering
    \includegraphics[width=0.95\linewidth]{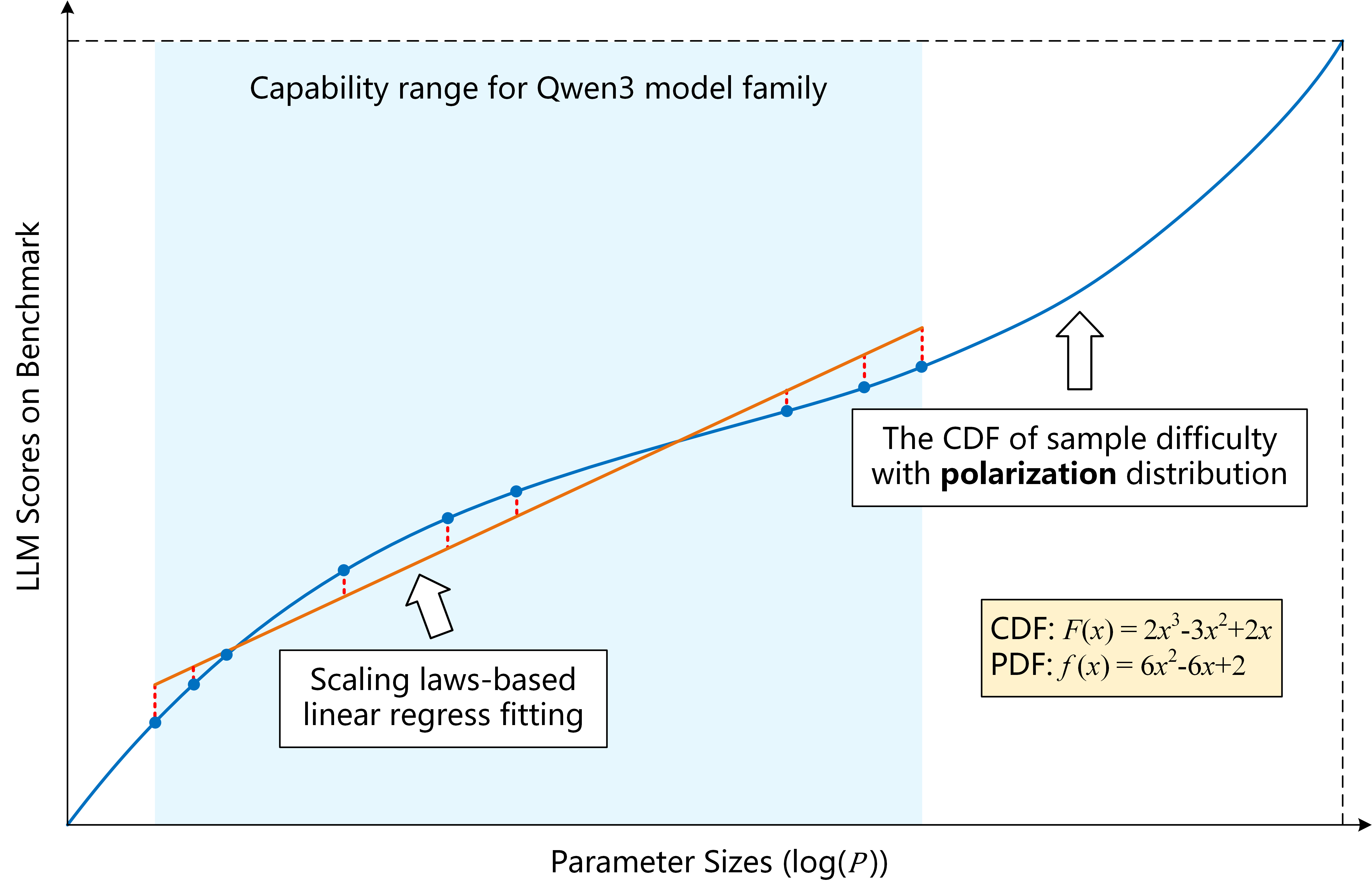}
    \caption{An intuitive example about sample difficulty polarization distribution}
    \label{fig:A2}
\end{figure}

\begin{table*}[t]
\centering
\begin{tabular}{lccccccc}
\toprule
log($P$)  & MMLU    & MMLU-Pro & SuperGPQA & GPQA & GSM8K & MATH \\ 
\midrule
log(235) & -5.7191 & -8.5774 & -2.9122 & -4.8463 & -8.4152  & -3.2004 \\
log(32)  & 1.8629  & 3.7220  & 2.7612  & 5.1861  & 1.4780   & -1.1720 \\
log(30)  & 0.0143  & 0.1556  & -0.9766 & -0.1046 & 0.2403   & -3.3555 \\
log(14)  & 4.1879  & 5.4061  & 1.3781  & -1.0818 & 5.0804   & 4.3064  \\
log(8)   & 3.3348  & 5.2992  & 1.5417  & 5.7070  & 5.4850   & 6.5242 \\
log(4)   & 3.5308  & 4.3428  & 1.7919  & 0.9225  & 7.2185   & 4.0822 \\
log(1.7) & -1.7729 & -3.0659 & -1.4466 & -4.2290 & -0.4609  & -1.2613 \\
log(0.6) & -5.4387 & -7.2825 & -2.1376 & -1.5538 & -10.6262 & -5.9236 \\ 
\bottomrule
\end{tabular}
\caption{Residual between observed benchmark scores and scaling law-based linear predictions across Qwen3 models.}
\label{table:A2}
\end{table*}

\begin{table*}[ht]
\centering
\begin{tabular}{cccccccc}
\toprule
Metrics   & MMLU    & MMLU-Pro & SuperGPQA & GPQA & GSM8K & MATH \\ 
\midrule
Std. Dev. & 3.9646  & 5.6827   & 2.1114    & 3.8782  & 6.4828  & 4.4293  \\
Skewness  & -0.5884 & -0.6288  & -0.1080   & 0.4300  & -0.7833 & 0.3147  \\
Kurtosis  & -1.3566 & -1.4465  & -1.8677   & -0.8630 & -0.6186 & -1.4252 \\
\bottomrule
\end{tabular}
\caption{Statistical analysis of residual distributions across six benchmarks under scaling law fitting}
\label{table:A3}
\end{table*}

\begin{figure*}[ht]
\centering
\subfigure[MMLU scores vs. Scaling laws fitting]{
  \includegraphics[width=0.32\linewidth]{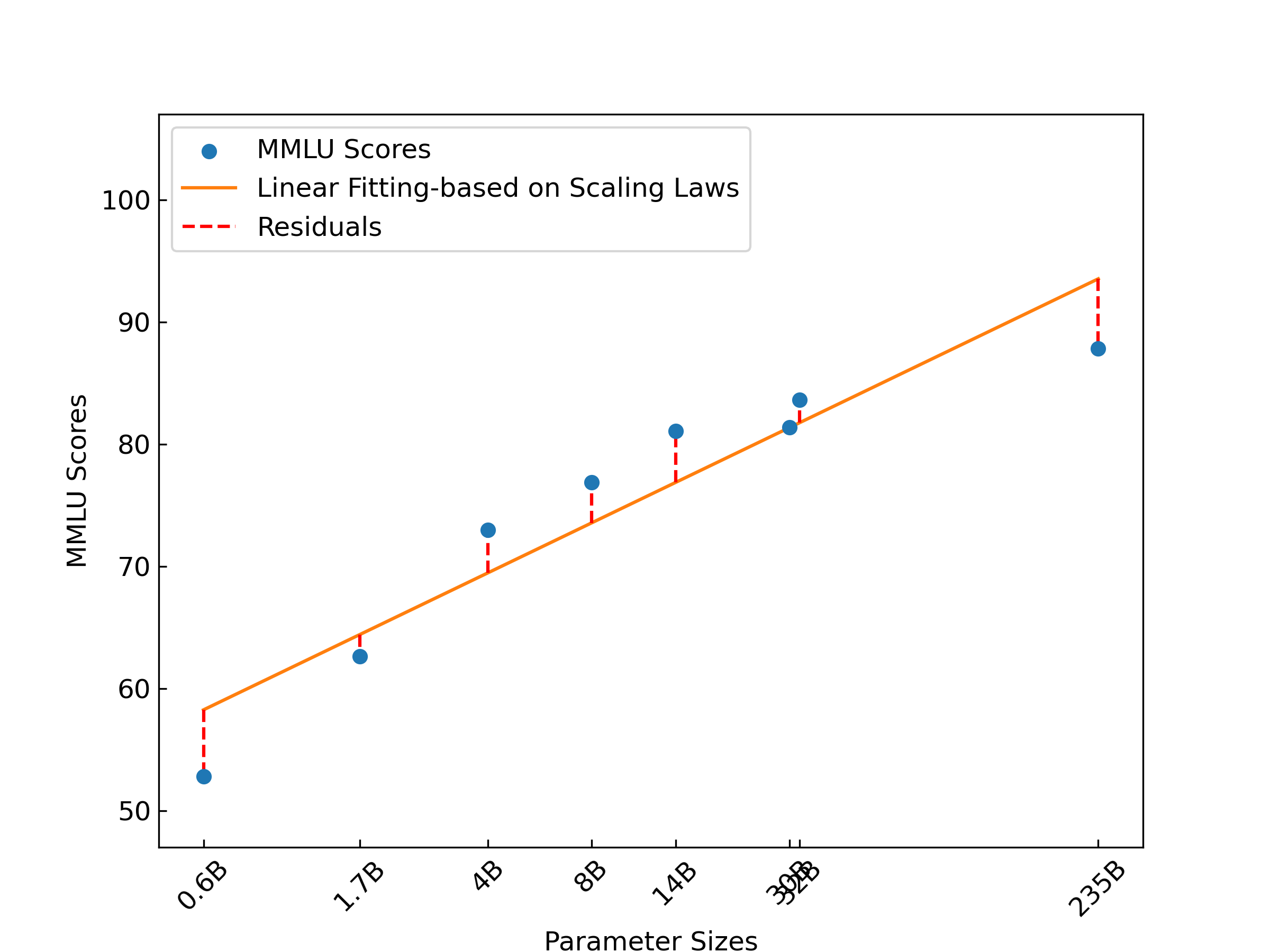}
  \label{fig:Aa}
}
\subfigure[MMLU-Pro scores vs. Scaling laws fitting]{
  \includegraphics[width=0.32\linewidth]{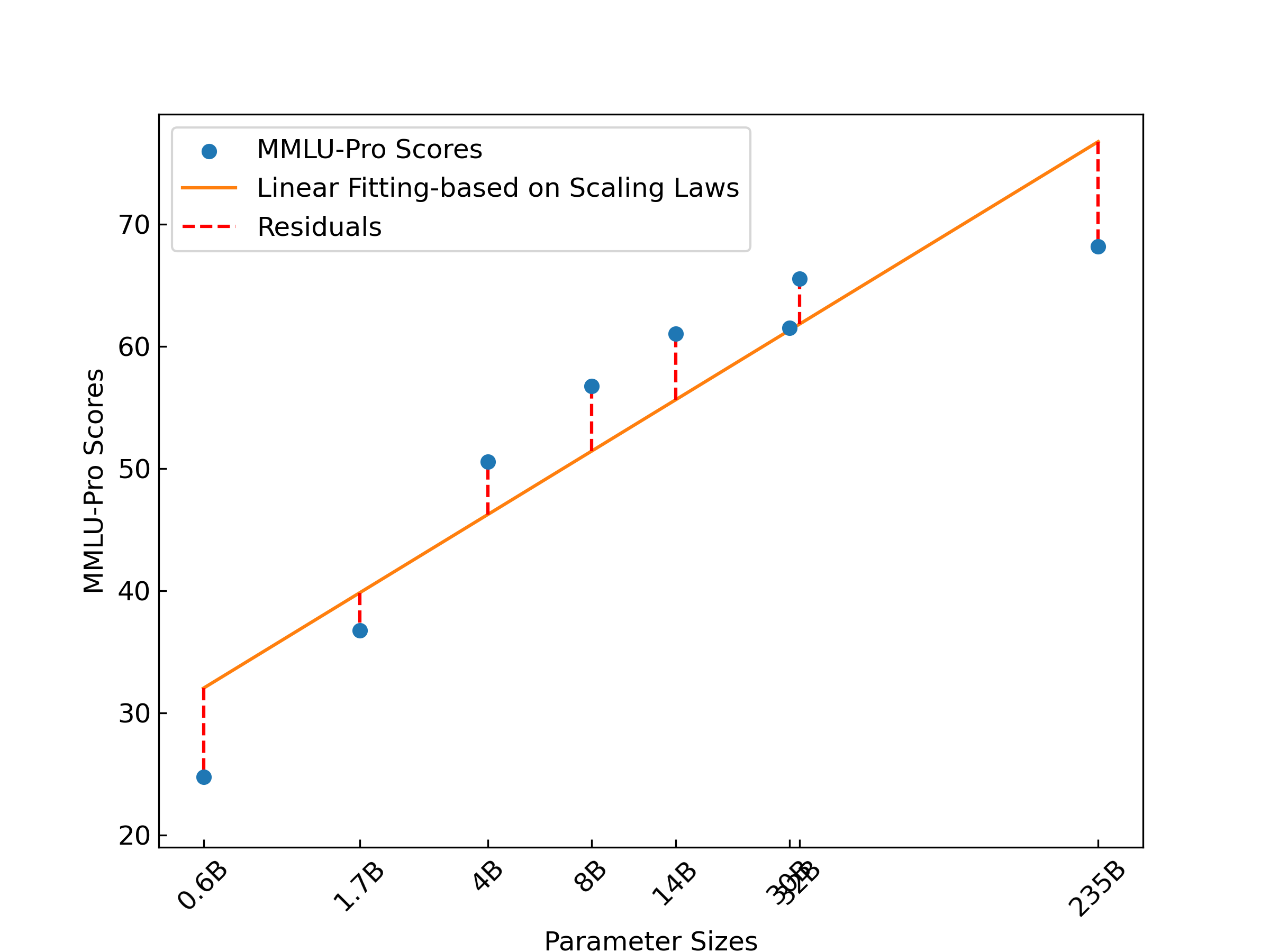}
  \label{fig:Ab}
}
\subfigure[GPQA scores vs. Scaling laws fitting]{
  \includegraphics[width=0.32\linewidth]{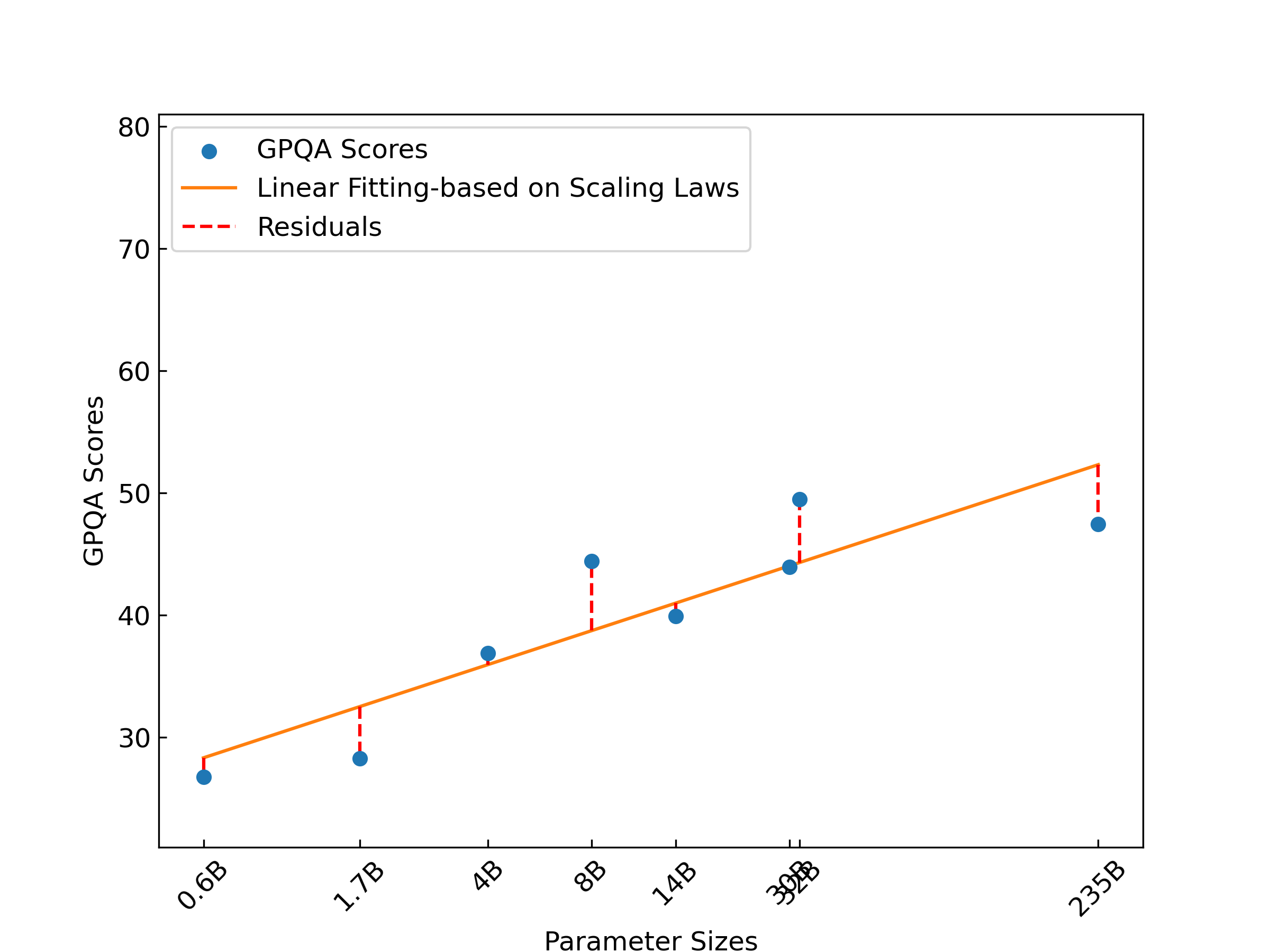}
  \label{fig:Ac}
}
\subfigure[SuperGPQA scores vs. Scaling laws fitting]{
  \includegraphics[width=0.32\linewidth]{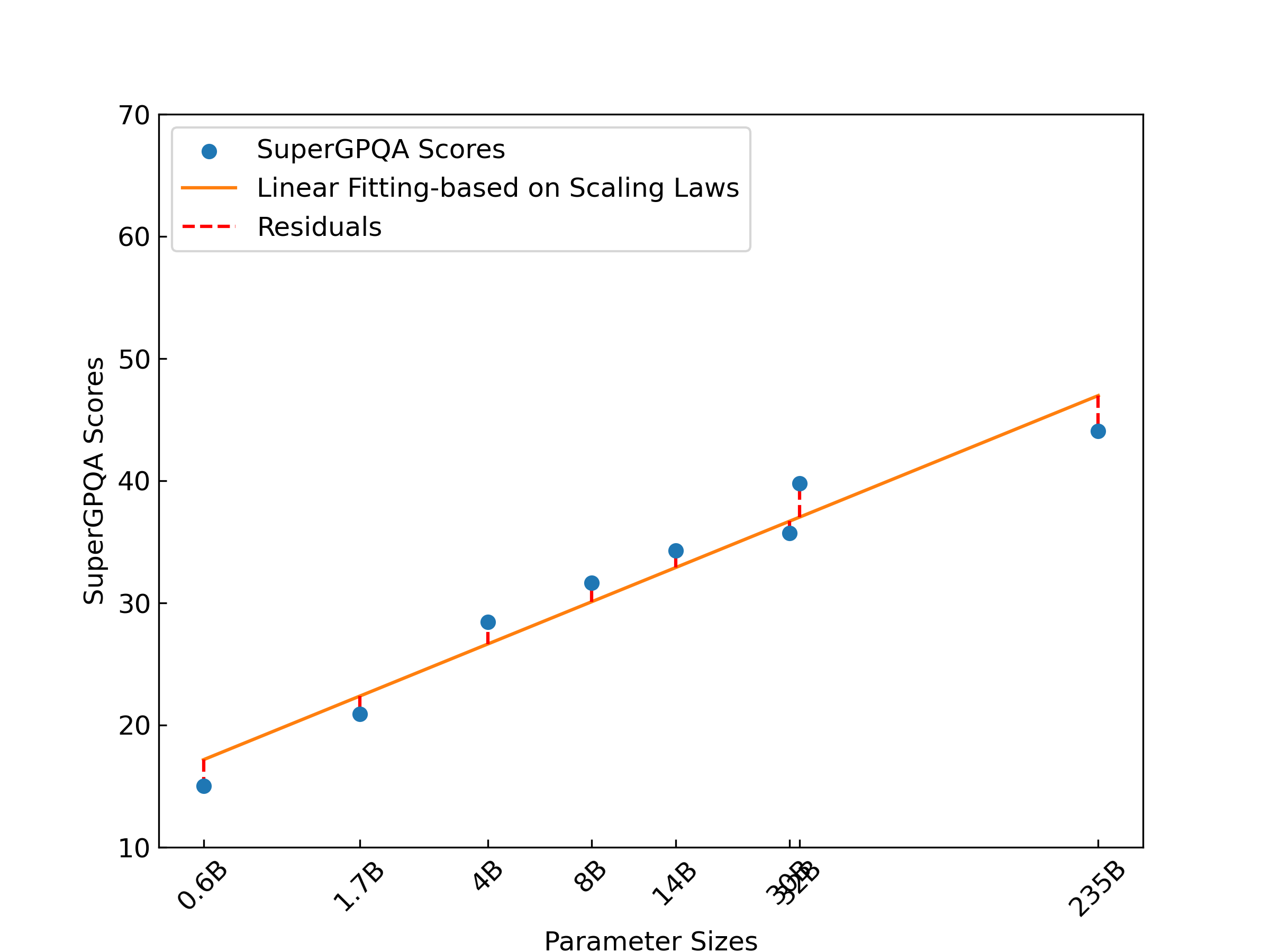}
  \label{fig:Ad}
}
\subfigure[GSM8K scores vs. Scaling laws fitting]{
  \includegraphics[width=0.32\linewidth]{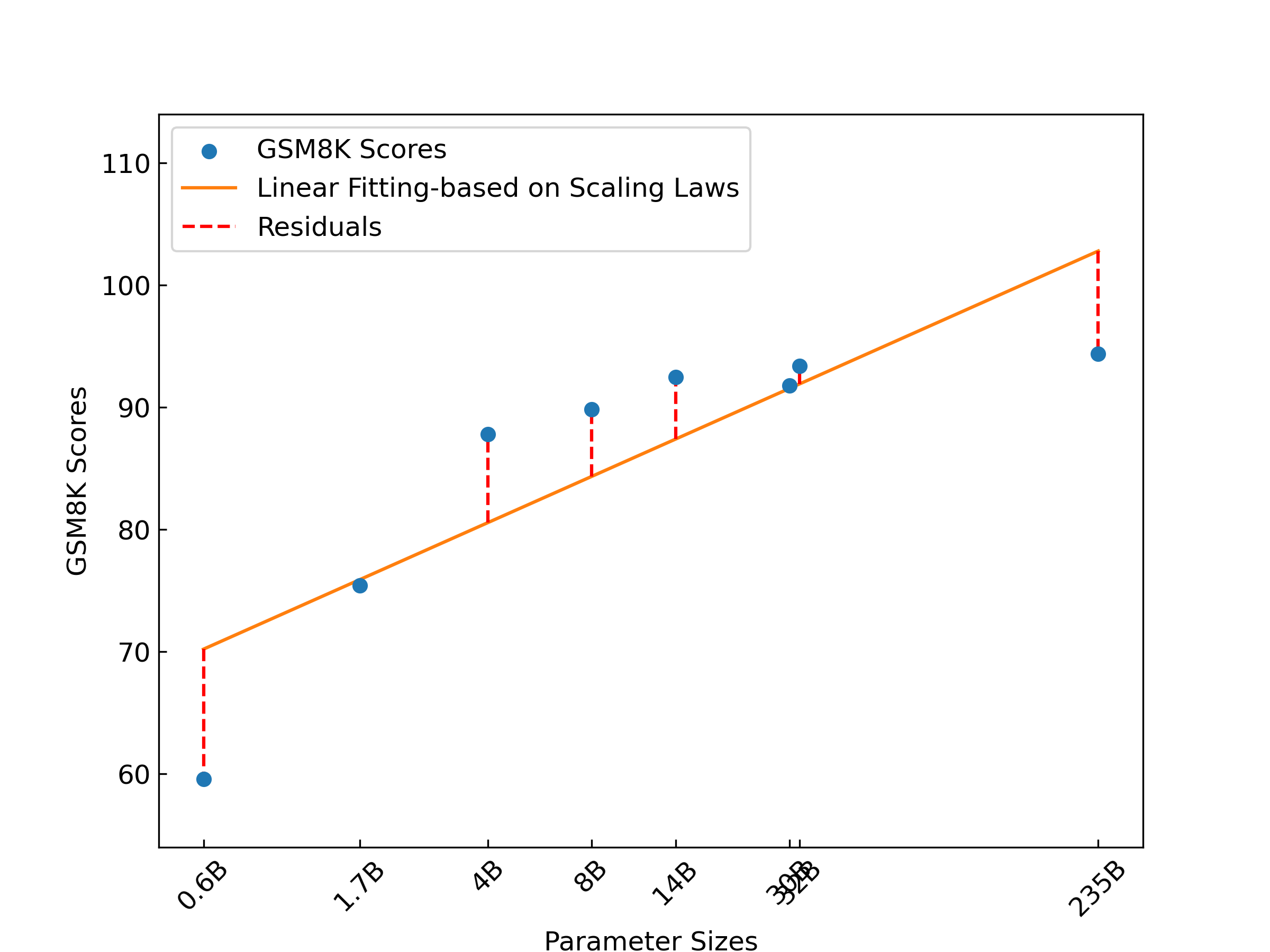}
  \label{fig:Ae}
}
\subfigure[MATH scores vs. Scaling laws fitting]{
  \includegraphics[width=0.32\linewidth]{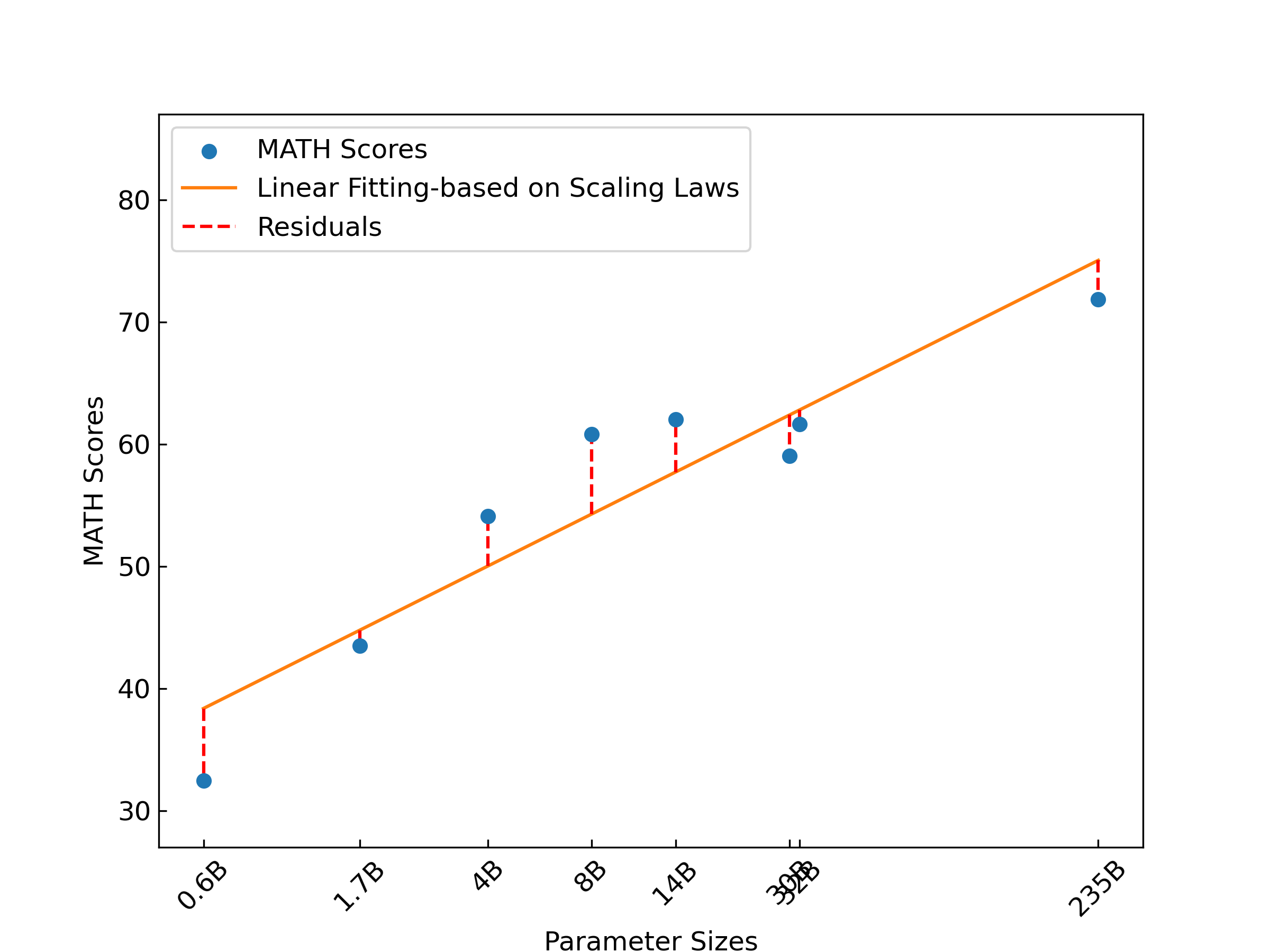}
  \label{fig:Af}
}
\caption{Linear fitting of benchmark performance across Qwen3 model family based on scaling laws}
\label{fig:A1}
\end{figure*}

The sample difficulty polarization indicates that the true performance curve as a function of model capability should be sigmoidal (S-shaped). An intuitive example is shown in Figure~\ref{fig:A2}. Performance initially accelerates as small models rapidly master the abundance of simple samples. It then enters a phase of maximum growth for medium-sized models tackling intermediate samples, before finally decelerating as large models hit a performance ceiling imposed by the difficult samples. When a linear regression is fitted to this S-shaped data, the model is forced to average out these changing growth rates. The steep slope of the curve's middle section inevitably "pulls up" the fitted line. As a mathematical consequence, this regression line will lie above the true performance data in the initial (accelerating) and final (decelerating) phases, and below the data in the middle (maximum growth) phase. This systemic mismatch perfectly explains the observed negative-positive-negative residual pattern ($\epsilon_i<0$ for small/large models, $\epsilon_i>0$ for medium-size models), providing powerful evidence of the structural bias within the benchmarks. 

To further validate this interpretation, a more comprehensive statistical analyses of the residuals was performed using \textbf{standard deviation}, \textbf{skewness} and \textbf{kurtosis} shown in Table~\ref{table:A3}. 

The standard deviation of residuals reflects alignment with the scaling law. SuperGPQA shows the smallest deviation, indicating strong alignment and a coherent difficulty structure. In contrast, GSM8K and MMLU-Pro exhibit large deviations of 6.48 and 5.68, reflecting substantial performance variance and weaker alignment. Skewness further reveals distributional bias: GSM8K's negative skew of -0.7833 indicates a long tail of difficult samples that depress overall model scores, while GPQA's positive skew of 0.4300 reflects a concentration of relatively simple items, leading to easier-than-expected scores. Finally, the consistently negative (platykurtic) kurtosis across benchmarks indicates flatter distributions with thinner tails, suggesting that deviations arise not from a few outliers but from a broad, systemic mismatch with scaling law expectations.

In summary, the above analysis reveals significant imbalance and structural bias in the difficulty distribution of these benchmarks. The preliminary finding point out that the difficulty distribution of the six benchmarks overall shows a trend of polarization. The subsequent residual analysis confirms this polarization, demonstrating how it systematically causes model performance to deviate from scaling law predictions. Such bias, characterized by high standard deviations, significant skewness and a systematically platykurtic residual distribution, limits the accuracy and effectiveness of benchmarks in reflecting true model generalization.


\begin{table*}[ht]
\centering
\begin{tabular}{lcccccc|c}
\toprule
\textbf{Benchmarks} & MMLU & MMLU-Pro & GPQA & SuperGPQA & GSM8K & MATH & \textbf{Score}\\ 
\midrule
Qwen3-235B-A22B & 83.13 & 60.82 & 45.06 & 39.69 & 95.53 & 87.04 & 70.04 \\
Qwen3-32B       & 79.57 & 57.10 & 42.41 & 37.03 & 95.34 & 79.72 & 66.54 \\
Qwen3-30B-A3B   & 76.57 & 53.26 & 42.63 & 34.42 & 94.72 & 77.16 & 64.26 \\
Qwen3-14B       & 76.59 & 51.86 & 40.40 & 32.93 & 94.75 & 75.29 & 63.14 \\
Qwen3-8B        & 71.94 & 48.60 & 37.28 & 29.73 & 93.99 & 81.36 & 61.75 \\
Qwen3-4B        & 67.08 & 42.61 & 32.37 & 26.00 & 89.01 & 80.64 & 57.53 \\
Qwen3-1.7B      & 56.17 & 31.56 & 31.47 & 20.31 & 78.87 & 76.96 & 49.97 \\
Qwen3-0.6B      & 45.23 & 24.03 & 33.48 & 16.22 & 32.89 & 16.98 & 27.83 \\ 
\midrule
LLaMA3-8B       & 61.37 & 34.22 & 32.37 & 20.78 & 80.78 & 21.77 & 41.97\\
GLM4-9B         & 73.31 & 42.46 & 33.04 & 22.51 & 91.62 & 58.91 & 54.64\\

\bottomrule
\end{tabular}
\caption{Full-benchmark evaluation scores of various LLMs across six benchmarks using zero-shot and non-CoT strategy.}
\label{tableB}
\end{table*}

\section{Prompting Strategy Underperforms on Mathematic Reasoning Tasks}
Few-shot prompting with chain-of-thought (CoT) reasoning has been widely regarded as a powerful technique for enhancing LLM performance on reasoning tasks. However, experimental results in Table~\ref{table1} and Table~\ref{tableB} consistently show that zero-shot without CoT prompting outperforms few-shot with CoT prompting on both GSM8K and MATH across multiple model scales. This observation challenges common assumptions There are three key factors contributing to this counterintuitive outcome:

\textbf{Error propagation in multi-step reasoning:} Mathematical problems typically involve a sequence of interdependent steps. CoT prompts encourage models to generate these intermediate steps explicitly. However, even minor arithmetic or logical errors in early steps often propagate, resulting in incorrect final answers. For example, there is a problem: “\textit{Alice has 3 boxes of pencils. Each box contains 12 pencils. She gives 10 pencils to her friend. How many pencils does she have left?}”. A CoT-based model may begin with “$3 \times 12 = 36$” correctly, but then mistakenly subtract 12 instead of 10, yielding “$36 - 12 = 24$” an incorrect final answer. In contrast, a zero-shot direct response model may internally compute all steps correctly without exposing intermediate vulnerabilities.

\textbf{Format drift and misalignment in few-shot prompts:} Few-shot CoT examples must be carefully constructed and aligned with the distribution and complexity of the target problems. In practice, slight discrepancies between the exemplars and the evaluation samples, such as differences in notation, problem structure, or reasoning depth, can lead to format drift or reasoning inconsistencies. Taking a example about polynomial factorization: “\textit{Simplify: $(3x + 2)(x - 5)$}”. The correct approach is straightforward algebraic expansion. However, if the few-shot examples involve lengthy step-by-step derivations of equations (e.g., solving quadratics using the quadratic formula), the model may mistakenly treat the expression as an equation to solve or add redundant steps such as $(3x + 2)(x - 5)=0\Rightarrow x=-\frac{2}{3}~and~x=5$. Such misalignment can confuse the model, causing it to either mimic suboptimal solution patterns or deviate from the intended reasoning format. In contrast, zero-shot prompting simply instructs the model to “simplify” the expression, prompting a direct and accurate response: “$3x^2-13x-10$”.

\textbf{Overthinking on simple task induced by CoT:} Modern LLMs trained with instruction tuning often possess strong zero-shot generalization abilities, especially on tasks with clear and deterministic solution paths. For relatively simple mathematical problems, introducing explicit reasoning steps through CoT can lead to unnecessary elaboration or semantic overreach, ultimately harming accuracy. The overthinking phenomenon occurs when the model attempts to simulate human-like reasoning even when direct computation suffices. For a simple task “\textit{A pencil costs \$2. How much do 6 pencils cost?}”, A zero-shot direct prompt typically elicits the correct answer “\$12” through implicit mental calculation. In contrast, CoT prompting might cause the model to generate a verbose sequence such as “\textit{Each pencil costs \$2. If we buy 6 pencils, we multiply 2 by 6. $2 \times 6 = 14$, so the answer is \$14.}” Not only does this introduce a calculation error, but it also illustrates how excessive reasoning steps can derail the model’s focus.

These findings underscore the importance of task- and model-specific prompting strategies. In particular, for reasoning-intensive benchmarks, zero-shot direct prompting can yield more stable and effective performance than the conventionally favored few-shot CoT approach.

\begin{table*}[ht]
\centering
\begin{tabular}{lcccccc|c}
\toprule
Benchmark & MMLU & MMLU-Pro & GPQA & SuperGPQA & GSM8K & MATH & \textbf{Score} \\ 
\midrule
LLaMA3-8B & $61.39_{\pm 13.61}$ & $33.11_{\pm 12.89}$ & $31.43_{\pm 10.43}$ & $20.91_{\pm 11.09}$ & $79.68_{\pm 7.68}$ & $22.08_{\pm 9.92}$ & $41.82_{\pm 11.07}$ \\
GLM4-9B   & $73.94_{\pm 11.94}$ & $43.08_{\pm 15.08}$ & $33.33_{\pm 13.67}$ & $22.90_{\pm 9.90 }$ & $90.72_{\pm 8.72}$ & $58.04_{\pm 14.04}$ & $54.98_{\pm 12.51}$ \\ \bottomrule
\end{tabular}\caption{Random sampling evaluation across six benchmarks using zero-shot and non-CoT strategy.}
\label{tableC}
\end{table*}

\section{Multi-scale Inference Prompts}

To ensure consistency and reduce variability caused by prompt design, all experiments in this study adopt standardized prompting templates across different benchmarks and models. The prompting strategy is unified as follows:

For MMLU, MMLU-Pro, GPQA, and SuperGPQA, the following instruction is used to elicit a single-choice answer: “\textit{There is a multiple-choice question with several options, and only one of the options is the correct answer. Please select the single correct option. The question is as follows. question: \textbf{ques\_desc}; options: \textbf{options\_text}. Please list the letter of your choice (e.g. A) in JSON format and output it with key ‘answer’ as the value. Do not explain your reasoning. Output JSON format: \text{\{‘answer’: ‘’\}}}”, where \textbf{ques\_desc} and \textbf{options\_text} are input variables.

For GSM8K and MATH, which are evaluated via automatic judgment of generated student solutions, the following verification prompt is used: “\textit{Here is a math problem with a standard answer and a student's solution. Please help me determine if the student's solution is correct.
question: \textbf{ques\_desc}; student solution: \textbf{stu\_solution}; standard solution: \textbf{std\_solution}; answer: \textbf{answer}. If the student's answer is correct, just output True; otherwise, just output False. No explanation is required. Please list your correct result (e.g. True or False) in JSON format and output it with key "result" as the value.  
Output JSON format: \text{\{‘result’: ‘’\}}}”, where \textbf{ques\_desc}, \textbf{stu\_solution}, \textbf{std\_solution} and \textbf{answer} are input variables.

These unified prompt templates ensure consistent model behavior across different scales and architectures, minimize instruction-following ambiguity, and facilitate reliable extraction of inference outcomes for comparative evaluation.

\end{document}